# Improving Tagging Consistency and Entity Coverage for Chemical Identification in Full-text Articles

Hyunjae Kim[1*], Mujeen Sung[1], Wonjin Yoon[1], Sungjoon Park[2], and Jaewoo Kang[1*]
[1]Korea University    [2]University of California, San Diego
*{hyunjae-kim, kangj}@korea.ac.kr

*Abstract*— This paper is a technical report on our system submitted to the chemical identification task of the BioCreative VII Track 2 challenge. The main feature of this challenge is that the data consists of full-text articles, while current datasets usually consist of only titles and abstracts. To effectively address the problem, we aim to improve tagging consistency and entity coverage using various methods such as majority voting within the same articles for named entity recognition (NER) and a hybrid approach that combines a dictionary and a neural model for normalization. In the experiments on the NLM-Chem dataset, we show that our methods improve models' performance, particularly in terms of recall. Finally, in the official evaluation of the challenge, our system was ranked 1st in NER by significantly outperforming the baseline model and more than 80 submissions from 16 teams.

## I. INTRODUCTION

Chemical identification, which involves finding chemical names in text and automatically linking them to the concepts in knowledge bases, is important for various downstream tasks such as drug-drug interaction extraction and document classification. Several datasets such as BC5CDR [1] were proposed to facilitate research on the chemical identification task. However, most current datasets consist of only the titles and abstracts of papers, despite the fact that other parts of papers contain more detailed and useful information. Recently, the NLM-Chem dataset [2] was proposed, which consists of full-text articles with chemical name and concept annotations. Using this data, BioCreative VII track 2 introduces a new challenge: chemical identification in full-text articles [3,4]. The task consists of two stages: (1) named entity recognition (NER) that involves predicting chemical entity boundaries in text, and (2) normalization that involves classifying the predicted entities into the corresponding biomedical concepts.

In this paper, we describe our final systems for the challenge and provide several experimental results and analyses on the NLM-Chem data. We consider two aspects to improve NER performance in full-text articles. First, models should consistently predict the same chemical entities within the same article. Unfortunately, models that use sentence-level input do not consistently predict the same entity as shown in Figure 1, i.e., they have the tagging inconsistency problem [5]. To alleviate this, recent works proposed to encode document-level information using complex modules such as an attention mechanism [5] and memory networks [6,7]. Instead, we use a simple post-processing method based on majority voting for aggregating model predictions in full text. Our method improves models' performance by improving tagging consistency.

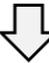

Fig. 1. The tagging inconsistency problem in sentence-level models and our majority voting-based solution. We highlight positive and negative predictions for the entity "FLLL32" in blue and red, respectively. Note that we deal with full-text articles in this work, but for simplicity we only include the title and abstract in this figure.

Second, different articles have very different distributions due to differences in research topics and writing styles. Consequently, unseen entities, which did not appear during training, often appear at the inference time. This tendency is more pronounced when the data is composed of full text than when it is composed of only titles and abstracts. We follow a

This research was supported by the MSIT (Ministry of Science and ICT), Korea, under the ICT Creative Consilience program (IITP-2021-2020-0-01819) supervised by the IITP (Institute for Information \& communications Technology Planning \& Evaluation) and National Research Foundation of Korea (NRF-2020R1A2C3010638, NRF-2014M3C9A3063541).

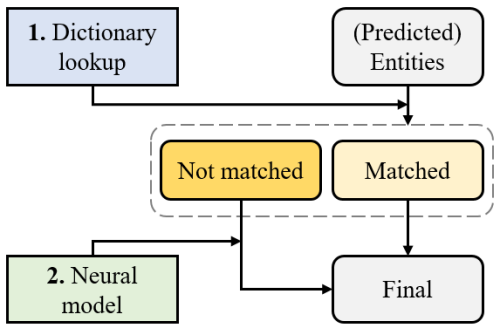

Fig. 2. The overview of the hybrid model. First, it performs dictionary lookup on the entities predicted by the NER model. Then, the neural model further normalizes entities not matched by the dictionary.

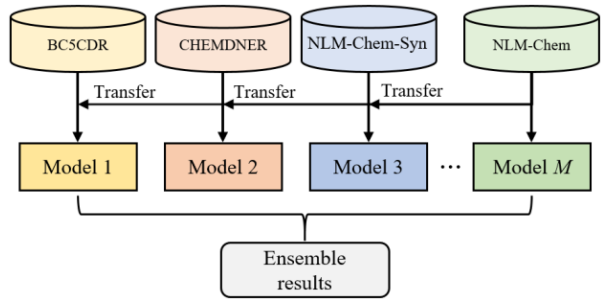

Fig. 3. The ensemble method to combine models trained on different training datasets.

current work [8] to measure the proportion of unseen entities. As a result, we found 48% of entities in the test set of NLM-Chem are unseen, whereas 35% are unseen in BC5CDR. This indicates that improving the generalization capabilities of models to unseen entities is important for NER at the full-text level. We use two current chemical NER datasets and an automatically-generated dataset by synonym replacement [9] as additional training resources. We pre-train models on the additional datasets, and then fine-tune the models on the NLM-Chem data. This transfer learning exposes models to more diverse chemical entities and contexts, improving entity coverage and the generalization to unseen entities. Also, we experiment with an ensemble method that combines different models trained on different datasets and find that it is more effective than combining models trained on the same data.

In normalization, we propose a hybrid model that combines a dictionary model and a neural model. Dictionary-based models usually achieve high precision but low recall due to the limited coverage of their dictionaries. On the other hand, neural network models achieve higher recall, but with less accuracy. We attempt to leverage the strengths of both while compensating for the weaknesses of each model. As shown in Figure 2, we first perform dictionary lookup, and then use a neural model to further predict entities that fail to be normalized by the dictionary model. This hybrid approach significantly improves entity coverage, which in turn improves normalization performance.

In sum, we use various methods such as majority voting and transfer learning for NER, and the hybrid approach for normalization. We show that our methods are simple but can boost NER and normalization performance by effectively improving tagging consistency and entity coverage. Finally, in the challenge, our system significantly outperformed the baseline model and was ranked 1st place in NER. However, in normalization, our system did not achieve satisfactory performance due to low precision, but we believe the hybrid approach is promising and can be improved in future work. The all submission results from 17 teams (including our team) can be found in [3].

## II. SYSTEM DESCRIPTION

Our system is a pipeline of an NER model and a normalization model, indicating that each model is independently trained, and they are combined at the inference time. Specifically, the NER model uses sentences as input and is trained by the sequence labeling objective. The normalization model uses the predictions of the NER model as input (i.e., predicted entities) and classifies them into the corresponding concepts.

### A. Transfer learning from different sources

In addition to NLM-Chem, we use two additional datasets BC5CDR [1] and CHEMDNER [10] to improve entity coverage and models' generalizability to unseen entities. We also generate the new synthetic data NLM-Chem-Syn by replacing entities in NLM-Chem with their synonyms, which are sampled from CTD (Comparative Toxicogenomics Database). NLM-Chem-Syn is 3x larger than the original NLM-Chem dataset. After the additional datasets are prepared, we pre-train NER models on the datasets, and then fine-tune the models with the NLM-Chem data. Note that we re-initialize the classification layer at the fine-tuning stage.

### B. Model ensemble

Ensemble methods theoretically reduce expected generalization errors by reducing the variance. They also have been shown empirically to be generally effective for many tasks and datasets. Thus, we adopt a simple ensemble method in our problem, which is based on majority voting. In addition, we test the effect of ensembling models with the same structure but different training data: NLM-Chem, BC5CDR, CHEMDNER, and NLM-Chem-Syn. Figure 3 depicts this ensemble method.

### C. Majority voting within the same articles

We simply address the tagging inconsistency problem using majority voting. First, we collect all inconsistent predictions in the same article. We then compute the majority for model predictions and change all the minority predictions to the majority. This majority voting method can be viewed as an ensemble of a single model's predictions within an article.

TABLE I. OUR FINAL SUBMISSION RESULTS IN THE BIOCREATIVE VII TRACK 2 CHALLENGE.

| Model | NER | | | Normalization | | |
|---|---|---|---|---|---|---|
| | P | R | F | P | R | F |
| Median | 0.848 | 0.814 | 0.837 | 0.712 | 0.776 | 0.775 |
| Baseline | 0.844 | 0.788 | 0.815 | **0.815** | 0.764 | **0.789** |
| **Ours 1** | 0.875 | <u>0.852</u> | <u>0.863</u> | 0.721 | 0.847 | 0.779 |
| **Ours 2** | **0.878** | 0.845 | 0.861 | <u>0.726</u> | **0.851** | <u>0.783</u> |
| **Ours 3** | <u>0.876</u> | **0.859** | **0.867** | 0.712 | <u>0.850</u> | 0.775 |

*We highlight the best scores in bold and underline the second best scores.
*P: precision, R: recall, F: F1 score

TABLE II. TOP 10 SUBMISSION RESULTS IN NER.

| Team / Run | P | R | F |
|---|---|---|---|
| **Our team / 3** | <u>0.8759</u> | 0.8587 | **0.8672** |
| **Our team / 1** | 0.8747 | 0.8523 | <u>0.8633</u> |
| **Our team / 2** | **0.8775** | 0.8447 | 0.8607 |
| 128 / 1 | 0.8544 | **0.8658** | 0.8600 |
| 143 / 1 | 0.8535 | 0.8608 | 0.8571 |
| 128 / 4 | 0.8457 | <u>0.8617</u> | 0.8536 |
| 128 / 2 | 0.8643 | 0.8403 | 0.8521 |
| 121 / 2 | 0.8461 | 0.8583 | 0.8521 |
| 121 / 1 | 0.8616 | 0.8415 | 0.8515 |
| 121 / 3 | 0.8580 | 0.8409 | 0.8494 |

*All teams are denoted as their unique indices. Our team is "Team 139."

### D. Handling sub-token entities

The NLM-Chem data has many *sub-token entities* that are sub-strings of a token rather than the whole string. For example, the token "Gly104Cys" has two sub-token entities "Gly" and "Cys." In the official evaluation of the challenge, models should predict sub-token entities, not the whole tokens. We found that sub-token entities mostly appear within mutation names, and about 90% of sub-token entities can be processed with simple regular expressions. Based on this, we perform post-processing on sub-token entities, which greatly improves performance in the official evaluation.

### E. Combining a dictionary and a neural model

We propose a hybrid approach to improve entity coverage in normalization. Specifically, we combine a dictionary lookup model and a neural network model. The dictionary model first performs normalization based on exact matching between entities and the dictionary. The neural model further performs the process on entities that are not matched by dictionary lookup.

### F. Implementation details

We use Bio-LM-large [11] as our NER backbone model. For the majority voting method, we only use entities that are longer than 2 and appear more than 40 times in the same article. The max length of input sequence is set to 512. We use the batch size of 24 and the learning rate of 1e-5. For final submission, 20 models were combined to create a "both" ensemble model. For normalization, we use the April 1st, 2021 version of the CTD as our chemical dictionary. We further expand the dictionary using mentions annotated in the training/development set of NLM-Chem and entities in Wikidata. We also use Ab3P to deal with abbreviations. For the neural model, we use BioSyn [12] with the SapBERT encoder [13] and use the same hyperparameters as the authors. We train the neural model on NLM-Chem. We also add the [CUI-LESS] embedding and normalize entities that do not have semantically similar names in the dictionary into "CUI-less."

## III. RESULTS AND ANALYSIS

### A. Overall results

Table I shows that the submission results for our top three models. In NER, our systems significantly outperformed the median (for 88 submissions from 17 teams) and the baseline model by achieving high performance in both precision and recall. Also, we show the top 10 submission results in NER in Table II. Our systems were ranked 1st, 2nd, and 3rd, respectively, indicating that our methods are consistently effective on different runs. For more information on the baseline and submission systems, please refer to [3].

Despite high performance in NER, our systems achieved similar or slightly better F1 scores compared to the median and perform below the baseline model in normalization. They achieved high recall, as we intended, but relatively with low precision. The cause of low precision may be due to error propagation from the NER models, i.e., false-positive predictions from the NER models may harm the precision of the normalization models. This could be alleviated by adopting the end-to-end approach [14]. Also, the normalization performance can be improved by designing high-precision dictionaries.

### B. Language model selection

To find the best-performing sentence encoder on the NLM-Chem data, we tested several variants of common pre-trained language models (PLMs) in the biomedical domain: BioBERT [15], PubMedBERT [16], and Bio-LM [11]. As a result, we found Bio-LM-large to be the most effective in our experiment (Table II). Also, we provide several observations from the experiment. First, although BioBERT usually works well on biomedical NLP tasks and achieves comparable (albeit slightly lower) performance with PubMedBERT and Bio-LM, it performed much worse on NLM-Chem. We conjecture differences in vocabulary may have had a significant impact on the performance. Second, PubMedBERT-full worked better than PubMedBERT. This indicates that pre-training on full-text articles may be effective for chemical NER at the full-text level. Third, Bio-LM-large performed better than Bio-LM-base, showing that model size can affect performance.

TABLE III. DIFFERENCES BETWEEN PRE-TRAINED LANGUAGE MODELS.

| Model | Vocab | Corpus | Size | P | R | F |
|---|---|---|---|---|---|---|
| BioBERT | Wiki+Books | Abstract | base | 0.833 | 0.863 | 0.848 |
| PubMedBERT | PubMed | Abstract | base | 0.862 | 0.882 | 0.872 |
| PubMedBERT-full | PubMed | Full text | base | 0.861 | 0.887 | 0.874 |
| Bio-LM-base | PubMed | Full text | base | 0.852 | **0.888** | 0.870 |
| Bio-LM-large | PubMed | Full text | large | **0.865** | 0.887 | **0.876** |

\*The performance was evaluated on the test set of NLM-Chem.
\*Abstract includes abstracts and also titles in this table.

TABLE IV. ABLATION STUDY IN NER.

|  | P | R | F |
|---|---|---|---|
| *Single model (fine-tune only)* | | | |
| NLM-Chem only | 0.865 | 0.887 | 0.876 |
| *Single model (transfer)* | | | |
| BC5CDR | 0.860 | 0.894 | 0.877 |
| CHEMDNER | 0.865 | 0.895 | 0.880 |
| NLM-Chem-Syn | 0.867 | 0.893 | 0.880 |
| *Ensemble* | | | |
| Fine-tune only | 0.868 | 0.892 | 0.879 |
| Transfer only | 0.872 | 0.899 | 0.885 |
| Both | 0.872 | 0.896 | 0.884 |
| *Ensemble (with majority voting)* | | | |
| Fine-tune only | 0.873 | 0.896 | 0.884 |
| Transfer only | 0.876 | **0.901** | 0.888 |
| Both | **0.880** | 0.898 | **0.889** |

\*The performance was evaluated on the test set of NLM-Chem.
\*For a description of models, please see the text.

TABLE V. ABLATION STUDY IN NORMALIZATION.

|  | P | R | F |
|---|---|---|---|
| Dictionary | **0.913** | 0.828 | 0.869 |
| Neural model | 0.841 | **0.885** | 0.862 |
| Hybrid model | 0.888 | 0.855 | **0.871** |

\*The performance was evaluated on the test set of NLM-Chem.
\*In this experiment, gold standard annotations are used as input.

*C. Effect of transfer learninig*

As shown in Table III, transfer learning improved models' performance by mostly improving entity coverage, i.e., recall. Although the synonym replacement method does not require additional labeling costs, it can be more effective than using existing human-labeled datasets.

*D. Effect of ensemble*

Table III shows that ensemble models outperform single models. Besides, we analyzed how the effect of ensembling varies according to the combinations of single models. We created three ensemble models, "Fine-tune only," "Transfer only," and "Both," which indicate the combination of models trained only with NLM-Chem, the combination of only transferred models, and the combination of both types of models, respectively. As a result, we found that ensembling models trained on different sources can be effective.

While we used the ensemble method for our final submission, the other methods such as transfer learning and majority voting are effective without ensembling. Especially, we recommend using majority voting if you do not have enough computational resources since the method is simple but consistently improved single models' performance by up to 1.0 F1 in our initial experiments on NLM-Chem.

*E. Effect of majority voting*

Table III shows that majority voting is simple but consistently improves performance. The method is particularly effective when there are many mentions of the same entity in one article, and there is severe tagging inconsistency. For instance, the article with PMID 2902420 has 137 mentions of the entity "FLLL32," and models predicted about 70% of the mentions as entities and the rest as not. In this case, the method corrected about 30% errors, which significantly improves performance.

*F. Effect of the hybrid model*

As shown in Table IV, dictionary lookup works very well in normalization if we have a high-quality dictionary. However, the method has low recall due to the limited coverage of the dictionary. Our hybrid model significantly improved recall, resulting in a higher F1 score. However, maintaining high precision of the dictionary model remains a challenge in future work.

IV. CONCLUSION

In this paper, we described our system for the chemical identification task of BioCreative VII Track 2. We focused on improving tagging consistency and entity coverage to perform chemical identification in full-text articles. To do so, we used various methods such as transfer learning, data augmentation with synoym replacement, ensemble of differently trained models, majority voting at the full-text level, and the hybrid approach that combines dictionary lookup and a neural model. In the experiments, we demonstrated the effectiveness of all methods that we used. Also, we had an important discussion regarding language model selection. We hope this work can provide insights into chemical identification in full-text articles.

REFERENCES

1. Li, J., Sun, Y., Johnson, R. J., Sciaky, D., Wei, C. H., Leaman, R., ... & Lu, Z. (2016). BioCreative V CDR task corpus: a resource for chemical disease relation extraction. Database, 2016.

2. Islamaj, R., Leaman, R., Kim, S., Kwon, D., Wei, C. H., Comeau, D.

C., ... & Lu, Z. (2021). NLM-Chem, a new resource for chemical entity recognition in PubMed full text literature. Scientific Data, 8(1), 1-12. Chicago

3. Leaman, R., Islamaj, R., Lu, Z. (2021). Overview of the NLM-Chem BioCreative VII track: Full-text Chemical Identification and Indexing in PubMed articles. Proceedings of the seventh BioCreative challenge evaluation workshop.

4. Islamaj, R., Leaman, R., Cissel, D., Cheng, M., Coss, C., Denicola, J., … & Lu, Z. (2021). The chemical corpus of the NLM-Chem BioCreative VII track: Full-text Chemical Identification and Indexing in PubMed articles. Proceedings of the seventh BioCreative challenge evaluation workshop.

5. Luo, L., Yang, Z., Yang, P., Zhang, Y., Wang, L., Lin, H., & Wang, J. (2018). An attention-based BiLSTM-CRF approach to document-level chemical named entity recognition. Bioinformatics, 34(8), 1381-1388.

6. Luo, Y., Xiao, F., & Zhao, H. (2020, April). Hierarchical contextualized representation for named entity recognition. In Proceedings of the AAAI Conference on Artificial Intelligence.

7. Gui, T., Ye, J., Zhang, Q., Zhou, Y., Gong, Y., & Huang, X. (2020, July). Leveraging Document-Level Label Consistency for Named Entity Recognition. In IJCAI.

8. Kim, H., & Kang, J. (2021). How Do Your Biomedical Named Entity Models Generalize to Novel Entities?. arXiv preprint arXiv:2101.00160.

9. Dai, X., & Adel, H. (2020). An Analysis of Simple Data Augmentation for Named Entity Recognition. In Proceedings of the 28th International Conference on Computational Linguistics.

10. Krallinger, M., Rabal, O., Leitner, F., Vazquez, M., Salgado, D., Lu, Z., ... & Valencia, A. (2015). The CHEMDNER corpus of chemicals and drugs and its annotation principles. Journal of cheminformatics, 7(1), 1-17.

11. Lewis, P., Ott, M., Du, J., & Stoyanov, V. (2020, November). Pretrained Language Models for Biomedical and Clinical Tasks: Understanding and Extending the State-of-the-Art. In Proceedings of the 3rd Clinical Natural Language Processing Workshop (pp. 146-157).

12. Sung, M., Jeon, H., Lee, J., & Kang, J. (2020). Biomedical Entity Representations with Synonym Marginalization. In Proceedings of the 58th Annual Meeting of the Association for Computational Linguistics.

13. Liu, F., Shareghi, E., Meng, Z., Basaldella, M., & Collier, N. (2021). Self-Alignment Pretraining for Biomedical Entity Representations. In Proceedings of the 2021 Conference of the North American Chapter of the Association for Computational Linguistics: Human Language Technologies.

14. Zhou, B., Cai, X., Zhang, Y., & Yuan, X. (2021, August). An End-to-End Progressive Multi-Task Learning Framework for Medical Named Entity Recognition and Normalization. In Proceedings of the 59th Annual Meeting of the Association for Computational Linguistics and the 11th International Joint Conference on Natural Language Processing.

15. Lee, J., Yoon, W., Kim, S., Kim, D., Kim, S., So, C. H., & Kang, J. (2020). BioBERT: a pre-trained biomedical language representation model for biomedical text mining. Bioinformatics, 36(4), 1234-1240.

16. Gu, Y., Tinn, R., Cheng, H., Lucas, M., Usuyama, N., Liu, X., ... & Poon, H. (2021). Domain-specific language model pretraining for biomedical natural language processing. ACM Transactions on Computing for Healthcare (HEALTH), 3(1), 1-23.